\documentclass[10pt, a4paper]{article}
\usepackage{lrec2022} \usepackage{multibib}
\newcites{languageresource}{Language Resources}
\usepackage[dvipdfmx]{graphicx}
\usepackage{tabularx}
\usepackage{soul}
\usepackage{hyperref}
\usepackage{xstring}
\usepackage{times}
\usepackage{latexsym}
\usepackage{breakurl}
\usepackage{amssymb}
\usepackage{amsfonts}
\usepackage{amsmath}
\usepackage{mathtools}
\usepackage{bm}
\usepackage{lscape}
\usepackage{algorithm,algpseudocode}
\usepackage{xcolor}
\usepackage{ulem}
\usepackage{caption}
\usepackage{setspace}
\usepackage{booktabs}
\usepackage{microtype}
\usepackage{here}

\usepackage{CJKutf8}
\usepackage[utf8]{inputenc}
\usepackage[T1]{fontenc}
\newcommand{\ja}[1]{\begin{CJK}{UTF8}{ipxm}#1\end{CJK}}

\usepackage{titlesec}
\titleformat{\section}{\normalfont\large\bfseries\center}{\thesection.}{1em}{}
\titleformat{\subsection}{\normalfont\SmallTitleFont\bfseries\raggedright}{\thesubsection.}{1em}{}
\titleformat{\subsubsection}{\normalfont\normalsize\bfseries\raggedright}{\thesubsubsection.}{1em}{}
\renewcommand\thesection{\arabic{section}}
\renewcommand\thesubsection{\thesection.\arabic{subsection}}
\renewcommand\thesubsubsection{\thesubsection.\arabic{subsubsection}}

\usepackage{epstopdf}
\usepackage[utf8]{inputenc}

\title{JParaCrawl v3.0: A Large-scale English-Japanese Parallel Corpus}

\name{Makoto Morishita, Katsuki Chousa, Jun Suzuki, and Masaaki Nagata}
\address{NTT Communication Science Laboratories, NTT Corporation\\
  2-4 Hikaridai, Seika-cho, Soraku-gun, Kyoto, 619-0237, Japan\\
  {\tt \{makoto.morishita.gr, katsuki.chousa.bg, masaaki.nagata.et\}@hco.ntt.co.jp}\\
  {\tt jun.suzuki@tohoku.ac.jp }}

\abstract{
Most current machine translation models are mainly trained with parallel corpora, and their translation accuracy largely depends on the quality and quantity of the corpora.
Although there are billions of parallel sentences for a few language pairs, effectively dealing with most language pairs is difficult due to a lack of publicly available parallel corpora.
This paper creates a large parallel corpus for English-Japanese, a language pair for which only limited resources are available, compared to such resource-rich languages as English-German.
It introduces a new web-based English-Japanese parallel corpus named JParaCrawl v3.0.
Our new corpus contains more than 21 million unique parallel sentence pairs, which is more than twice as many as the previous JParaCrawl v2.0 corpus.
Through experiments, we empirically show how our new corpus boosts the accuracy of machine translation models on various domains.
The JParaCrawl v3.0 corpus will eventually be publicly available online for research purposes.
\\ \newline
\Keywords{parallel corpus, machine translation, English, Japanese}
}

\begin{document}

\maketitleabstract

\section{Introduction}
The current neural machine translation models are generally trained by supervised approaches~\cite{sutskever14sequencetosequence,bahdanau15alignandtranslate,luong15emnlp,vaswani17transformer}, denoting reliance on parallel corpora.
However, since publicly available parallel corpora remain limited, training a model for many language pairs is difficult.
Thus, constructing a parallel corpus is crucial for expanding the applicability of machine translation.

This paper introduces a new large-scale web-based parallel corpus for English-Japanese for which only limited parallel corpora are available.
One of the current largest parallel corpora for this language pair is JParaCrawl~\cite{morishita20lrec}, which is constructed by crawling the web and automatically aligning parallel sentences.
However, this corpus contains around 10 million sentence pairs, which is still limited compared to the other resource-rich language pairs, and it is somewhat outdated because it was created two years ago.
We entirely re-crawled the web to update the corpus and applied a different approach to extract parallel sentences.
We collected more than 21 million unique sentence pairs, which is more than twice as many as the previous JParaCrawl corpus.
We experimentally show how the new crawled corpus increases the accuracy of machine translation for English-Japanese and Japanese-English.
Our new corpus, named JParaCrawl v3.0, will be publicly available through our website\footnote{\url{http://www.kecl.ntt.co.jp/icl/lirg/jparacrawl/}} for further researches.

Our contributions can be summarized:
\begin{itemize}
    \item We constructed a large-scale English-Japanese parallel corpus, which contains more than 21 million sentence pairs, on top of the previous JParaCrawl corpus.

    \item We empirically confirmed that our corpus boosted the accuracy of English-Japanese and Japanese-English machine translation in broad domains.

    \item We plan to release our new parallel corpus for further researches.
\end{itemize}

\section{Related Work}
\label{sec:related_work}
There are several sources for creating parallel corpora.
One typical source is the parallel documents written by international organizations.
An example is Europarl~\cite{koehn05europarl}, which was created from the proceedings of the European Parliament.
\newcite{ziemski16un} complied the United Nations parallel corpus from the translated documents of the UN.
Professional translators usually translate these texts, which sometimes contain such meta-data as document IDs that allow easy alignment of them.
Unfortunately, since these parallel documents are not commonly available,
these corpora are limited to a few language pairs and narrow domains.

Another critical source is the web.
Many websites are written in several languages, and parallel sentences can be extracted from them.
Thus the web is a fruitful source for creating a large parallel corpus in many languages and broader domains.
In an earlier work, \newcite{uszkoreit10mining} created a large-scale distributed system to mine parallel sentences from the web and books.
\newcite{smith13dirtcheap} proposed a method to mine parallel sentences from Common Crawl\footnote{\url{https://commoncrawl.org/}}, a free web crawl archive.
Recently, some works created large parallel corpora from Wikipedia or Common Crawl~\cite{schwenk19wikimatrix,schwenk19ccmatrix} with the latest parallel sentence alignment method, which uses multilingual sentence embeddings.

ParaCrawl is an important work that creates a large-scale parallel corpus for 24 European languages from the web~\cite{banon20paracrawl}.
Since it continuously updates the corpus, it continues to grow.
Inspired by that work, \newcite{morishita20lrec} created a web-based large-scale parallel corpus for English-Japanese, where no large parallel corpus was available.
Their corpus, called JParaCrawl, amassed more than 10 million sentences and is the largest publicly available parallel corpus for that language pair.
However, the current JParaCrawl corpus remains tiny compared to the other resource-rich language pairs, and thus its translation accuracy is inferior to other resource-rich languages.
Thus a larger parallel corpus must be created for English-Japanese.
In this work, we extend the JParaCrawl corpus by re-crawling the web and applying a new parallel sentence extraction method, as described in the following section.

\section{JParaCrawl v3.0}
\label{sec:jparacrawl}

\begin{table}[tbp]
\centering
\small
\begin{tabular}{lrr}
\toprule
\textbf{Version} & \textbf{\# sentences} & \textbf{\# words}\\ \midrule
v1.0 & $4,817,172$ & $125,216,523$ \\
v2.0 & $8,809,771$ & $234,393,978$ \\
v3.0 & $21,481,513$ & $502,445,763$ \\
\bottomrule
\end{tabular}
\caption{Number of unique sentence pairs and words on English side in JParaCrawl corpus. In this work, we created version 3.0. }
\label{tab:corpus_stats}
\end{table}

\begin{table*}[!tbp]
\centering
\small
\begin{tabular}{llrr}
\toprule
\textbf{Test set} & \textbf{Domain} & \textbf{\# sentences} & \textbf{\# words}\\ \midrule
ASPEC & Scientific Papers & $1,812$ & $39,573$\\
JESC & Movie Subtitles & $2,000$ & $13,617$\\
KFTT & Wikipedia Articles & $1,160$ & $22,063$\\
TED (tst2015)  & TED Talks & $1,194$ & $20,367$\\
Business Scene Dialogue Corpus & Dialogues & $2,120$ & $19,619$ \\
WMT20 News En-Ja  & News & $1,000$ & $22,141$ \\
WMT20 News Ja-En  & News & $993$ & $24,423$ \\
WMT21 News En-Ja  & News & $1,000$ & $23,305$ \\
WMT21 News Ja-En  & News & $1,005$ & $24,771$ \\
WMT19 Robustness En-Ja (MTNT2019) & Reddit & $1,392$ & $19,988$ \\
WMT19 Robustness Ja-En (MTNT2019) & Reddit & $1,111$ & $13,390$ \\
WMT20 Robustness Set1 En-Ja  & Wikipedia Comments & $1,100$ & $29,419$ \\
WMT20 Robustness Set2 En-Ja  & Reddit & $1,376$ & $20,011$ \\
WMT20 Robustness Set2 Ja-En  & Reddit & $997$ & $15,866$ \\
IWSLT21 Simultaneous Translation En-Ja Dev & TED Talks & $1,442$ & $20,677$ \\
\bottomrule
\end{tabular}
\caption{Number of sentences and words on English side in test sets}\label{tab:test_set_corpus_stats}
\end{table*}

\begin{table}[!tb]
\centering
\small
\begin{tabular}{lrr}
\toprule
\textbf{Data} & \textbf{\# sentences} & \textbf{\# words}\\ \midrule
ASPEC & $3,008,500$ & $68,929,413$ \\
JESC & $2,797,388$ & $19,339,040$ \\
KFTT & $440,288$ & $9,737,715$ \\
TED & $223,108$ & $3,877,868$ \\
\bottomrule
\end{tabular}
\caption{Number of sentences and words on English side in training sets. Original version of ASPEC contains 3.0 million sentences, but we used only first 2.0 million for training based on previous work \protect\cite{neubig14wat}.}\label{tab:train_set_corpus_stats}
\end{table}

\begin{table}[htb]
  \centering
  \small
  \tabcolsep=1pt
  \begin{tabular}{lp{40mm}}
  \toprule
    \multicolumn{2}{c}{\textbf{Common Settings}} \\ \midrule
    Architecture            &  Transformer~\cite{vaswani17transformer} \\
    Enc-Dec layers & 6 \\
    Optimizer              &   Adam ($\beta_{1}=0.9, \beta_{2}=0.98, \epsilon=1\times10^{-8}$)~\cite{kingma14adam}  \\
    Learning rate schedule &   Inverse square root decay     \\
    Warmup steps           &   4,000  \\
    Max learning rate      &   0.001  \\
    Dropout                &   0.3~\cite{srivastava14dropout}  \\
    Gradient clipping      &   1.0~\cite{pascanu13clipping}  \\
    Label smoothing        &   $\epsilon_{ls}=0.1$~\cite{szegedy16cvpr}     \\
    Mini-batch size        &   512,000 tokens~\cite{ott18scaling}\\
    Number of updates      &   36,000 steps (v3.0), 24,000 steps (v1,0, v2.0) \\
    Averaging              &   Save checkpoint for every 100 steps and take an average of last 8 checkpoints \\
    Beam size              &   6 with length normalization~\cite{wu16gnmt}\\
    \midrule
    \multicolumn{2}{c}{\textbf{Small Settings}} \\ \midrule
    Attention heads & 4 \\
    Word-embedding dimension & 512 \\
    Feed-forward dimension & 1,024 \\
    \midrule
    \multicolumn{2}{c}{\textbf{Base Settings}} \\ \midrule
    Attention heads & 8 \\
    Word-embedding dimension & 512 \\
    Feed-forward dimension & 2,048 \\
    \midrule
    \multicolumn{2}{c}{\textbf{Big Settings}} \\ \midrule
    Attention heads & 16 \\
    Word-embedding dimension & 1,024\\
    Feed-forward dimension & 4,096\\
  \bottomrule
  \end{tabular}
  \caption{List of hyperparameters}
  \label{tab:hyper-parameter}
\end{table}

\begin{table*}[!tb]
\centering
\small
\begin{tabular}{lrrrrrrrrrr}
\toprule
 & \multicolumn{4}{c}{\bf English-to-Japanese} & \multicolumn{4}{c}{\bf Japanese-to-English}\\
\cmidrule(lr){2-5}
\cmidrule(lr){6-9}
& & \multicolumn{3}{c}{\bf JParaCrawl} & & \multicolumn{3}{c}{\bf JParaCrawl} \\
\cmidrule(lr){3-5}
\cmidrule(lr){7-9}
{\bf Test set} & {\bf In-domain} & {\bf v1.0} & {\bf v2.0} & {\bf v3.0} & {\bf In-domain} & {\bf v1.0} & {\bf v2.0} & {\bf v3.0} \\ \midrule
ASPEC & $44.3$ & $24.7$ & $26.5$ & $\bf 26.8$ & $28.7$ & $18.3$ & $19.7$ & $\bf 20.8$ \\
JESC  & $14.5$ & $\bf 6.6$ & $6.5$ & $6.5$ & $17.8$ & $7.0$ & $7.5$ & $\bf 8.4$ \\
KFTT  & $31.8$ & $17.1$ & $\bf 18.9$ & $18.1$ & $23.4$ & $13.7$ & $16.2$ & $\bf 17.0$ \\
TED (tst2015) & $11.1$ & $11.5$ & $12.6$ & $\bf 13.1$ & $13.7$ & $11.0$ & $11.9$ & $\bf 12.0$ \\
Business Scene Dialogue Corpus & --- & $12.4$ & $13.5$ & $\bf 13.9$ & --- & $17.4$ & $19.6$ & $\bf 19.9$ \\
WMT20 News En-Ja  & --- & $20.7$ & $21.9$ & $\bf 23.5$ & --- & $21.3$ & $23.3$ & $\bf 23.9$ \\
WMT20 News Ja-En  & --- & $20.1$ & $22.8$ & $\bf 23.5$ & --- & $19.2$ & $21.0$ & $\bf 21.9$ \\
WMT21 News En-Ja  & --- & $21.1$ & $21.8$ & $\bf 25.0$ & --- & $21.9$ & $23.1$ & $\bf 24.3$ \\
WMT21 News Ja-En  & --- & $19.6$ & $21.5$ & $\bf 22.4$ & --- & $18.1$ & $20.7$ & $\bf 21.3$ \\
WMT19 Robustness En-Ja (MTNT2019) & --- & $12.4$ & $12.5$ & $\bf 14.4$ & --- & $15.6$ & $16.8$ & $\bf 17.3$ \\
WMT19 Robustness Ja-En (MTNT2019)  & --- & $11.5$ & $12.3$ & $\bf 12.8$ & --- & $16.0$ & $17.2$ & $\bf 17.7$ \\
WMT20 Robustness Set1 En-Ja  & --- & $15.2$ & $15.8$ & $\bf 18.7$ & --- & $20.0$ & $20.6$ & $\bf 21.6$ \\
WMT20 Robustness Set2 En-Ja  & --- & $12.7$ & $13.0$ & $\bf 14.8$ & --- & $16.4$ & $17.4$ & $\bf 17.9$ \\
WMT20 Robustness Set2 Ja-En  & --- & $7.9$ & $8.2$ & $\bf 8.6$ & --- & $12.0$ & $12.6$ & $\bf 14.0$ \\
IWSLT21 Simultaneous Translation En-Ja Dev & --- & $12.5$ & $13.3$ & $\bf 14.5$ & --- & $12.9$ & $14.3$ & $\bf 14.5$ \\
\bottomrule
\end{tabular}
\caption{BLEU scores of models trained with in-domain training set, JParaCrawl v1.0, v2.0, and v3.0. Best scores among JParaCrawl models are highlighted in bold.}\label{tab:experimental_results}
\end{table*}

This paper extends the current JParaCrawl v2.0 corpus by further crawling the web and extracting parallel sentences.
Our methods are based on previous ParaCrawl and JParaCrawl projects~\cite{banon20paracrawl,morishita20lrec}.
We described the detailed process in the following sections.

\subsection{Find Websites Written in Parallel}
Our method extracts parallel sentences from the web.
Thus, the first step is finding a website that has parallel sentences.
This method is based on the hypothesis that websites containing the same English and Japanese sentences might have parallel texts.
To list such parallel websites, we analyzed all the Common Crawl text archive data released from March 2019 to August 2021\footnote{During this period, the Common Crawl project released 25 archives, and their text size was about 212 TB.}.
We identified the language in the archive by {\tt CLD2}\footnote{\url{https://github.com/CLD2Owners/cld2}} and listed 100,000 large websites that roughly have the same size of English and Japanese texts.
For this step, we used {\tt extractor}\footnote{\url{https://github.com/paracrawl/extractor}} that was provided by the ParaCrawl project.

We ignored the data released before March 2019 because they were already analyzed by the previous JParaCrawl project and focused more on the latest Common Crawl archive.
We checked the website lists generated by this procedure and found that 70\% were not listed in the previous JParaCrawl~\cite{morishita20lrec}.

\subsection{Crawl the Found Websites}
Next we crawled the websites listed in the previous step
with {\tt Heritrix}\footnote{\url{https://github.com/internetarchive/heritrix3}} at most 48 hours for each one.
Although the previous JParaCrawl only focused on plain texts, in this work, we also crawled PDF and Microsoft Word documents in addition to plain text to
extract more parallel sentences because the Japanese government and companies sometimes release their news on PDFs.

\subsection{Extract Parallel Sentences}
Next we extracted parallel sentences from the crawled archives with
 {\tt Bitextor}\footnote{\url{https://github.com/bitextor/bitextor}} provided by the ParaCrawl project.
We added Japanese support on Bitextor 8 and used it for this work.
For parallel document and sentence alignment, we used machine a translation-based alignment toolkit {\tt bleualign}~\cite{sennrich11bleualign}.
It first translates the Japanese sentences into English with a machine translation system and finds an English-Japanese sentence pair that maximizes the BLEU scores~\cite{papineni02bleu}.
For the Japanese-English translations, we used a Transformer-based neural machine translation (NMT) model trained with JParaCrawl v2.0.
Preliminary experiments found that {\tt bleualign} outperformed a bilingual lexicon-based method\footnote{The previous JParaCrawl used {\tt hunalign}~\cite{varga05hunalign}, which relies on a bilingual lexicon.}.

\subsection{Filter Out Noisy Sentences}
As the last step, we filtered out the incorrectly aligned or poorly translated noisy sentence pairs with the
{\tt Bicleaner}\footnote{\url{https://github.com/bitextor/bicleaner}} toolkit~\cite{prompsit18bicleaner}.
Then we concatenated the clean parallel sentences and JParaCrawl v2.0 and deduplicated them.
From these steps, we created a new large JParaCrawl v3.0 that contains more than 21 million sentences, which is more than twice as many as the previous JParaCrawl v2.0.

Table~\ref{tab:corpus_stats} shows the number of unique sentence pairs in the previous and the new JParaCrawl v3.0 corpus.
Note that this number is different from our previous paper~\cite{morishita20lrec}, because here we are reporting the number of {\it unique} sentence pairs.

\begin{figure*}[tb]
\centering
\small
\begin{tabular}{ll}
\toprule
Source & \ja{院内に「\uwave{濃厚接触者}」はいませんが、接触者全員にPCR検査を実施し、} \\
       & \ja{女性が関係した病棟などを閉鎖して徹底的に消毒するということです。}\\ \midrule
Reference & There are no known ``\uwave{close contacts}'' in the hospital, but all contacts will be subjected to PCR tests, \\
          & and the wards and other areas where the women had been will be closed and thoroughly disinfected. \\ \midrule
JParaCrawl v1.0 & There is no ``\uwave{strong contact person}'' in the hospital, but a PCR test will be conducted for all the contacts, \\
                & and women will close the wards and thoroughly disinfect them. \\ \midrule
JParaCrawl v2.0 & Although there is no ``\uwave{strong contact person}'' in the hospital, PCR tests will be performed on all contact\\
                & persons, and the wards related to women will be closed and thoroughly disinfected. \\ \midrule
JParaCrawl v3.0 & There are no ``\uwave{close contacts}'' in the hospital, but PCR tests will be conducted for all contacts, \\
                & and the wards related to women will be closed and thoroughly disinfected. \\
\bottomrule
\end{tabular}
\caption{Example translations of trained models. Example is from WMT21 News Ja-En test set.}\label{fig:example_translation}
\end{figure*}

\section{Experiments}
\subsection{Experimental Settings}

As an experiment, we trained an NMT model with JParaCrawl v3.0 and evaluated its accuracy on various test sets to confirm the effect of our new collected corpus.

\subsubsection{Test Sets}

To evaluate the NMT models on various domains, we tested our models on the 15 test sets listed in Table~\ref{tab:test_set_corpus_stats}.
We used all the test sets in our previous work~\cite{morishita20lrec},
which
included the Asian Scientific Paper Excerpt Corpus (ASPEC)~\cite{nakazawa16aspec}, the Japanese-English Subtitle Corpus (JESC)~\cite{pryzant17jesc}, the Kyoto Free Translation Task (KFTT)~\cite{neubig11kftt}, and TED talks (tst2015)~\cite{cettolo12wit3}.
We also evaluated our models on the Business Scene Dialogue Corpus~\cite{rikters19bsd} to check whether they worked on conversations.
We also added test sets from shared tasks: WMT 2020, 2021 news translation shared tasks~\cite{barrault20wmt,akhbardeh21wmt}, WMT 2019, 2020 robustness shared tasks~\cite{wmt19robustness,wmt20robustness}, and the IWSLT 2021 simultaneous translation task~\cite{anastasopoulos21iwslt}.
Although some of the test sets are intended for specific translation directions (e.g., En$\rightarrow$Ja), we used them for both En$\rightarrow$Ja and Ja$\rightarrow$En directions for reference.

Some corpora have an in-domain training set, as shown in Table~\ref{tab:train_set_corpus_stats}.
For comparison, we trained our model with these training sets and reported the BLEU scores.

\subsubsection{Training Settings}
First, we tokenized the corpus into sub-words with the {\tt sentencepiece} toolkit with a vocabulary size of 32,000~\cite{kudo18sentencepiece}.
Then we trained the NMT models with the {\tt fairseq} toolkit~\cite{ott19fairseq}.
Our models are based on Transformer~\cite{vaswani17transformer} and trained with three settings: small, base, and large.
Table~\ref{tab:hyper-parameter} shows the detailed training settings.
We used the small model for TED (tst2015), the base model for KFTT, and the big model for the others.
Note that these settings are almost the same as our previous work~\cite{morishita20lrec} for a fair comparison, except we changed the number of updates for the v3.0 models because the new corpus is too large to converge in 24,000 steps.
We evaluated the model with the {\tt sacreBLEU} toolkit~\cite{post18sacrebleu} and reported the BLEU scores~\cite{papineni02bleu}.
For the evaluations, we NFKC-normalized all the test sets for consistency with our previous experiments~\cite{morishita20lrec}.

\subsection{Experimental Result}
Table~\ref{tab:experimental_results} shows the BLEU scores on various test sets.
Our corpus is not designed as a specific domain but as a general one.
Thus, unsurprisingly, the JParaCrawl model did not reach the model's score trained with in-domain data.
The model trained with JParaCrawl v3.0 achieved the best score on all the test sets on Japanese-English and 13 of 15 test sets on English-Japanese.
These results clearly show that our new parallel corpus increased the accuracy of the NMT models on various domains, including scientific papers, news, and dialogues.

Our v3.0 model worked especially well on the WMT21 news translation tasks.
We believe that this is because the previous JParaCrawl v2.0 was based on the web in 2019, and so it might not have included terms frequently used in 2021.
For example, news articles in 2021 cited words related to COVID-19, a term that was not obviously less frequently used in 2019.
Perhaps we need to continue to update the parallel corpus to reflect the latest terms.

\subsection{Translation Example}
Figure~\ref{fig:example_translation} shows an example translation of JParaCrawl v1.0, v2.0, and v3.0.
We chose this example from the WMT21 news translation test set because it is related to COVID-19.
This input includes the Japanese phrase ``\ja{濃厚接触者}'', which should have been translated to ``close contacts.'' But the v1.0 and v2.0 models incorrectly translated the language to ``strong contact person.''
In contrast, the model trained with v3.0 correctly translated the phrase to ``close contacts.''
Similar to this example, we identified many improvements in the articles related to COVID-19.
These results support our hypothesis that our model trained with the v3.0 corpus correctly translated the terms and language frequently used in recent years.

\section{Conclusion}
This paper introduced an updated version of the large English-Japanese parallel corpus called JParaCrawl.
We re-crawled parallel websites by analyzing the latest CommonCrawl archive and extended the crawl target to PDF and Word documents.
After filtering out noisy sentences, the new JParaCrawl v3.0 included more than 21 million unique sentence pairs.
We empirically confirmed that the new corpus boosts the translation accuracy on various domains, especially on the trendiest news articles.
Our future work will update the JParaCrawl corpus and propose better alignment/filtering techniques.
The new JParaCrawl v3.0 will be available on our website for further research.
We expect that JParaCrawl v3.0 will support future research and products.

\section{Acknowledgements}
We gratefully acknowledge the ParaCrawl project for releasing its software and its members with whom we engaged in valuable discussions.

\section{Bibliographical References}\label{reference}

\bibliographystyle{lrec2022-bib}
\bibliography{myplain,main}

\end{document}